\documentclass{bmvc2k}
\usepackage{xcolor}
\usepackage{soul}
\usepackage[utf8]{inputenc}
\usepackage{helvet}
\usepackage{courier}

\newcommand{\tabincell}[2]{\begin{tabular}{@{}#1@{}}#2\end{tabular}}

\usepackage{graphicx}
\usepackage{float}

\newfloat{figtab}{htb}{fgtb}
\makeatletter
\newcommand\figcaption{\def\@captype{figure}\caption}
\newcommand\tabcaption{\def\@captype{table}\caption}
\makeatother

\title{Pixel-level Semantics Guided \\ Image Colorization}

\addauthor{Jiaojiao Zhao}{j.zhao3@uva.nl}{1}
\addauthor{Li Liu}{liuli1213@gmail.com}{2}
\addauthor{Cees G.M. Snoek}{cgmsnoek@uva.nl}{1}
\addauthor{Jungong Han}{jungonghan77@gmail.com}{3}
\addauthor{Ling Shao}{ling.shao@ieee.org}{2}

\addinstitution{
 University of Amsterdam, \\ The Netherlands\\
}
\addinstitution{
 Inception Institute of Artificial Intelligence,\\ UAE\\
}
\addinstitution{
 Lancaster University, \\ UK\\
}

\runninghead{ZHAO, LIU, SNOEK, HAN, SHAO}{IMAGE COLORIZATION}

\begin{document}

\maketitle

\begin{abstract}
While many image colorization algorithms have recently shown the capability of producing plausible color versions from gray-scale photographs, they still suffer from the problems of context confusion and edge color bleeding. To address context confusion, we propose to incorporate the pixel-level object semantics to guide the image colorization. The rationale is that human beings perceive and distinguish colors based on the object's semantic categories. We propose a hierarchical neural network with two branches. One branch learns what the object is while the other branch learns the object's colors. The network jointly optimizes a semantic segmentation loss and a colorization loss. To attack edge color bleeding we generate more continuous color maps with sharp edges by adopting a joint bilateral upsamping layer at inference. Our network is trained on PASCAL VOC2012 and COCO-stuff with semantic segmentation labels and it produces more realistic and finer results compared to the colorization state-of-the-art.  
\end{abstract}

\begin{figure}
	
	\centering
	\includegraphics[width=0.95\linewidth]{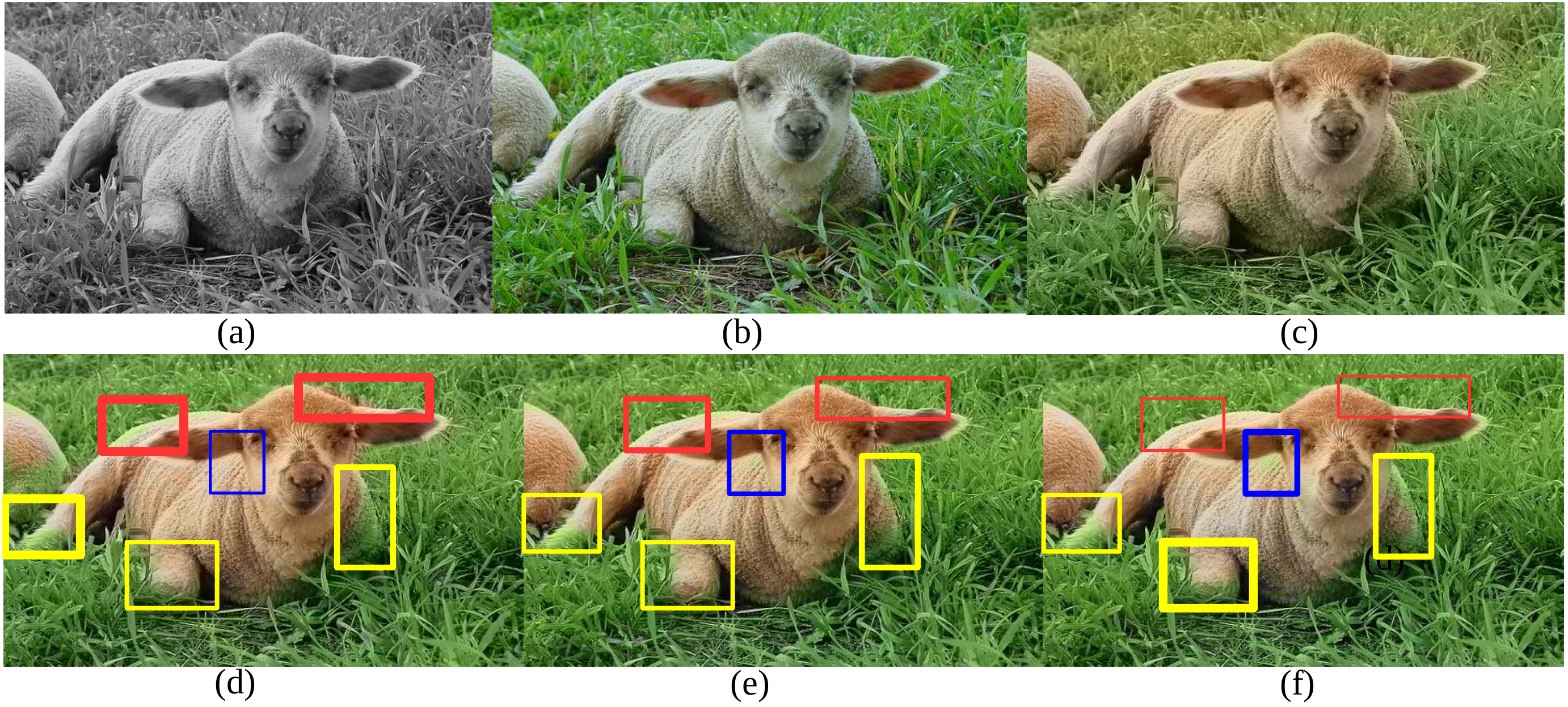}

	\caption[small] {Colorization results generated using the same input image but at different scales. (a) gray-scale image (300x500); (b) ground-truth (300x500); (c) proposed colorized version of the input image at size 224x224; colorized versions of~\cite{Zhang2016} with the input images at sizes:  (d) 176x176; (e) 224x224; (f) 300x300. We highlight some parts for edge bleeding (red boxes) and context confusion (blue boxes) to compare the results with various scales. For all boxes, the thicker the linewidth, the worse the result. Furthermore, even for the difficult parts (yellow boxes), the proposed method has better colorization abilities. Best viewed in color on zoomed-in screen.} 		
	\label{fig: Fig.1}
	
\end{figure}

\section{Introduction}
\label{sec:intro}
Colorizing a gray-scale image~\cite{Charpiat2008,Mor2009,Cheng2015,Isola2017,Cao2017,Gua2017} has wide applications in a variety of computer vision tasks, such as image compression~\cite{Baig2017}, outline and cartoon creations ~\cite{Frans2017,Qu2006}, and infrared images ~\cite{Limmer2016} and remote sensing images colorizations~\cite{Guo2017}. Human beings excel in assigning colors to gray-scale images as they can easily recognize the objects and have gained knowledge about their colors. No one doubts the sea is typically blue and a dog is naturally never green. Certainly, lots of objects have diverse colors which makes the prediction quite subjective. However, it remains a big challenge for machines to acquire both the world knowledge and ``imagination'' that humans possess. Previous works require reference images~\cite{Gupta2012,Liu2008} or color scribbles~\cite{Levin2004} as guidance. Recently, several automatic approaches ~\cite{Zhang2016,Larsson2016,Iizuka2016,Royer2017,Zhang2017} were proposed based on deep convolutional neural networks. Despite the improved colorization, there are still common pitfalls that make the colorized images appear less realistic. For example, color confusion in objects caused by incorrect semantic understanding and boundary color bleeding by scale variation. Our objective is to effectively address both problems to generate better colorized images with high quality.

Both traditional~\cite{Chia2011,Irony2005} and recent colorization solutions~\cite{Zhang2016,Larsson2016,Iizuka2016} have highlighted the importance of semantics. In~\cite{Zhang2016,Zhang2017}, Zhang \textit{et al.} apply cross-channel encoding as self-supervised feature learning with semantic interpretability. In~\cite{Larsson2016}, Larsson \textit{et al.} claim that interpreting the semantic composition of the scene and localizing objects are key to colorizing arbitrary images. Larsson \textit{et al.} pre-train a network on ImageNet for a classification task which provides the global semantic supervision. Iizuka \textit{et al.}~\cite{Iizuka2016} leverage a large-scale scene classification database to train a model, exploiting the class-labels of the dataset to learn the global priors. These works only explore the image-level classification semantics. As it is stated in~\cite{Jifeng2016}, the image-level classification task favors translation invariance. Obviously, colorization task needs representations that are translation-variant to an extent. From this perspective, semantic segmentation task which also needs translation-variant representations is more reasonable to provide pixel-level semantics for colorization. Deep CNNs have shown great successes on semantic segmentation ~\cite{She2016,Chen16}, especially with deconvolutional layers ~\cite{Noh2015}. It gives a class label to each pixel. Similarly, referring to ~\cite{Zhang2016,Larsson2016}, colorization assigns each pixel a color distribution. Both challenges can be viewed as an image-to-image prediction problem and formulated as a pixel-wise classification task. Our proposed network is able to harmoniously train with two loss functions of semantic segmentation and colorization. 

Edge color persistence is another common problem for existing colorization methods~\cite{Huang2005,Luan2007,Zhang2016,Larsson2016}. To speed up training and reduce memory consumption, deep convolutional neural networks prefer to take small fixed-sized images as inputs. However, test images may be at any scale compared to the resized training images. In this case, the well-trained model imposes a conflict between the semantics and edge color persistence on test images. In Figure~\ref{fig: Fig.1}, we present some results produced by~\cite{Zhang2016} when the input image is at different scales. The smaller the scale of the input image, the better understanding of the object colors but the worse the edge colors. Moreover, the downsampling and upsampling operations in the networks also cause edge color blurring. Inspired by the idea of joint bilateral filtering~\cite{KCLU07} which keeps edges clear and sharp, we propose a joint bilateral upsampling (JBU) layer for producing more continuous color maps of the same size with the original gray-scale image. 

Our contributions include: \textbf{(1)} We propose a multi-task convolutional neural network to learn what the object is and the colors of the object should be. \textbf{(2)} We propose a joint bilateral upsampling layer to generate a color inference from a color distribution. The method produces realistic color images with sharp edges. \textbf{(3)} The two strategies can be embedded in many existing colorization networks.

\begin{figure}[t]
	\begin{center}
		\includegraphics[width=0.95\linewidth]{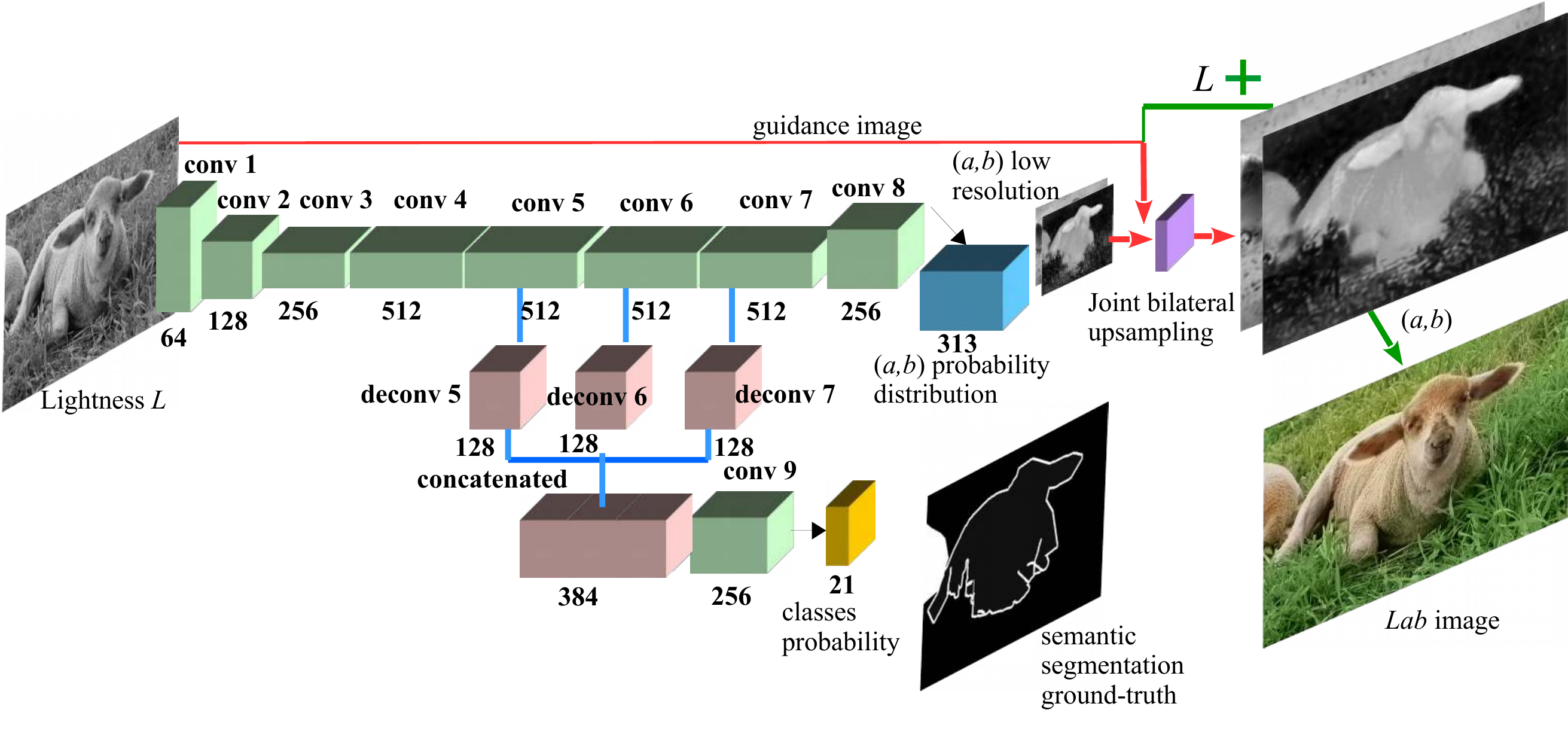}
	\end{center}
	\vspace{-8mm}
	\caption [small]{Our hierarchical network structure includes semantic segmentation and colorization. The semantic branch learns the pixel-wise object classes for the gray-scale images, which acts as a coarse classification. The colorization branch performs a finer classification according to the learned semantics. We apply multipath deconvolutional layers to improve semantic segmentation. At inference, a joint bilateral upsamping layer is added for predicting the colors. }
	\label{fig: Fig.2}
	
\end{figure}

\section{Methodology}
\label{sec:method}

We propose a hierarchical architecture to jointly optimize semantic segmentation and colorization. In order to estimate a specific color for each pixel from a color distribution, we propose a joint bilateral upsampling layer at the test phase. The architecture is illustrated in Figure~\ref{fig: Fig.2} and detailed next.

\subsection{Loss Function with Semantic Priors}

We consider the CIE \textit{Lab} color space to perform the colorization task as only two channels \textit{a} and \textit{b} need to be learned. The lightness channel \textit{L} with a height $H$ and a width $W$ is defined as an input $ \mathbf{X} \in \mathbf{R}^{H\times{W}\times{1}} $ and the output $ \mathbf{\hat{Y}} \in \mathbf{R}^{H\times{W}\times{2}}$ represents the two color channels \textit{a}, \textit{b}. The colorization problem is to learn a mapping function $f: \mathbf{X}\rightarrow\mathbf{\hat{Y}}$. Following the work in~\cite{Zhang2016}, we divide the color  \textit{a}\textit{b} space into $Q=313$ bins where $Q$ is the number of discrete \textit{a}\textit{b} values. The deep neural network shown in Figure~\ref{fig: Fig.2} is constructed to encode $\mathbf{\hat{Z}}=G(\mathbf{X})$ to a probability distribution over possible colors $\mathbf{\hat{Z}} \in [0,1]^{{H}\times{W}\times{Q}}$. Given a ground-truth $\mathbf{Z}$, a multinomial cross entropy loss function with class-rebalance for colorization $L_c$ is formulated as:
\begin{equation}
\label{eqn:Lc}
L_c(\mathbf{\hat{Z}},\mathbf{Z})=-\sum_{h,w}{v_c(\mathbf{Z}_{h,w})\sum_q{\mathbf{Z}_{h,w,q}log(\mathbf{\hat{Z}}_{h,w,q})}},
\end{equation} 
where $v_c(\cdot)$ indicates the weights for rebalancing the loss based on color-class rarity.

We jointly learn the other loss function $L_s$ specifically for semantic segmentation. Generally, semantic segmentation should be performed in the RGB image domain due to that colors are important for semantic understanding. However, the input of our network is a gray-scale image which is more difficult to segment. Fortunately, the network incorporating colorization learning supplies color information which in turn strengthens the semantic segmentation for gray-scale images. The mutual benefit between the two learning parts is the core of our network. Actually, semantic segmentation, as a supplementary means for colorization, is not required to be very precise. We define a weighted cross entropy loss with the standard softmax function $\mathbf{E}$ for semantic segmentation as:
\begin{equation}
\label{eqn:Ls}
L_s(\mathbf{X})=-\sum_{h,w}{v_s(\mathbf{X}_{h,w})log(\mathbf{E}(\mathbf{X}_{h,w};\theta))},
\end{equation}  
where $v_s(\cdot)$ is the weighting terms to rebalance the loss based on object-category rarity.

Finally, our loss function $L$ is a combination of $L_c$ and $L_s$ and can be jointly optimized:
\begin{equation}
\label{eqn:L}
L={\lambda_c}L_c+{\lambda_s}L_s,
\end{equation}
where $\lambda_i$ is the weights to balance the losses for colorization and semantic segmentation.

\subsection{Inference by Joint Bilateral Upsampling Layer}
There are some strategies for a point estimation from a color distribution according to~\cite{Zhang2016,Larsson2016}. Usually, taking the mode of the prediction distribution for each pixel will provide a vibrant result but with splotches. Alternatively, applying the mean of the distribution will produce desaturated results. In order to find a balance between the two factors, Zhang \textit{et al.} in~\cite{Zhang2016} propose to use the annealed-mean of the distribution. The trick helps to achieve more acceptable results but cannot predict the edge colors well. We draw inspiration from the joint bilateral filter~\cite{KCLU07} to address the issue. The joint bilateral filter uses both a spatial filter kernel on the initially generated color maps and a range filter kernel on a second guidance image (the gray-scale image here) to estimate the color values. More formally, for one position $p$, the filtered result on the color channel $c$ $(= a , b)$ is:
\begin{equation}
\label{eqn:J}
{J^c}_p=\frac{1}{k_p}\sum_{q\in\Omega}{{I^c}_qf(\Vert{p-q}\Vert)g(\Vert{\tilde{I_p}}-\tilde{I_q}\Vert)},
\end{equation} 
where $f$ is the spatial filter kernel, e.g., a Gaussian filter, and $g$ is the range filter kernel, centered at the gray-scale image ($\tilde{I}$) intensity value at $p$. $\Omega$ is the spatial support of the kernel $f$, and $k_p$ is a normalizing factor. Edges are preserved since the bilateral filter $f\cdot g$ takes on smaller values as the range distance and/or the spatial distances increase. Thus, the strategy hits three birds with one stone. That is, it decreases the splotches, keeps the colors saturated and makes the edges sharp and clear.

The input size of the network is usually small to speed-up training and reduce memory consumption so the outputs are low-resolution color maps. In order to get a colorized version at any test image resolution but with fine edges, we further adopt the joint bilateral upsampling (JBU) method. Let
$p$ and $q$ denote (integer) coordinates of pixels in the gray-scale image $\tilde{I}$, and $p_\downarrow$ and $q_\downarrow$ denote the corresponding (possibly fractional) coordinates in the low resolution output
$S^c$, the upsampled solution $\tilde{S}^c$ is:
\begin{equation}
\label{eqn:S}
{\tilde{S}^c}_p=\frac{1}{k_p}\sum_{q_\downarrow \in\Omega}{{S^c}_{q_\downarrow}f(\Vert{p_\downarrow-q_\downarrow}\Vert)g(\Vert{\tilde{I_p}}-\tilde{I_q}\Vert)},
\end{equation} 
We implement the joint bilateral upsampling using a neural network layer resulting in an end-to-end solution at inference.

\subsection{Network Architecture}
Our hierarchical network structure is specifically shown in Figure~\ref{fig: Fig.2}. The bottom layers conv1-conv4 are shared by the two tasks for learning the low level features. The high-level features contain more semantic information. We add three deconvolutional layers respectively after the top layers conv5, conv6 and conv7. Then the feature maps from the deconvolutional layers are concatenated for semantic segmentation, which is appropriate to capturing the fine-details of an object. Intuitively, the network will firstly recognize the object and then assign colors to the object. At the training phase we jointly learn the two tasks and at the test phase a joint bilateral upsampling layer is added to produce the final results.

\section{Experiments}
\label{sec:exper}

\subsection{Experimental Settings}

\noindent\textbf{Datesets:} Two datasets including the PASCAL VOC2012~\cite{pascal2012} and the COCO-stuff~\cite{Ca2016} are used. The former one is a common semantic segmentation dataset with 20 object classes and a background class. Our experiments are performed on the 10582 images for training and the 1449 images in validation set for testing. The COCO-stuff is a subset of the COCO dataset~\cite{Lin2014} generated for scene parsing, containing 182 object classes and a background class on 9000 training images and 1000 test images. Each input image is rescaled to 224x224.

\noindent\textbf{Implementation details:} Commonly available pixel-level annotations intended for semantic segmentation are sufficient for our method to improve colorization. We don’t need new pixel-level annotations for colorization. We train our network with joint semantic segmentation and colorization losses with the weights $\lambda_s:\lambda_c=100:1$ so that the two losses are similar in magnitude. Our multi-task learning for simultaneously optimizing colorization and semantic segmentation effectively avoids overfitting. 40 epochs are trained for the PASCAL VOC2012 and 20 epochs for the COCO-stuff. A single epoch takes approximately 20 minutes on a GTX Titan X GPU. The run time for the network is about 25 ms per image and the model size is 147.5 Mb, which are a little worse than those of ~\cite{Zhang2016} (22 ms and 128.9 Mb), as we have a few more layers for semantic segmentation. They are much better than ~\cite{Larsson2016} with a run time of 225 ms and a model size of 561 Mb and ~\cite{Iizuka2016} with a run time of 80 ms and a model size of 694.7 Mb. When performing JBU, the domain parameter $\sigma_s$ for the spatial Gaussian kernel is set to 3 and the range parameter $\sigma_r$ for the intensity kernel is set to 15.

\subsection{Illustration of Reasonability of the Strategies}
A simple experiment is performed for stressing that colors are critical for semantic segmentation. We apply the Deeplab-ResNet101 model~\cite{Chen16} trained on the PASCAL VOC2012 training set for semantic segmentation and test on three versions of the validation images, including gray-scale images, original color images and our colorized images. The mean intersection over union (IoU) is adopted to evaluate the segmentation results. As seen in Figure~\ref{fig: Fig.3}, with the original color information, the performance 86$\%$ is much better than that of the gray images 79$\%$. The performance of our proposed colorized images is 5$\%$ lower than that of the original RGB images. One reason is the imperfection of the generated colors. We believe another reason is that the model was trained on the original color images. However, as we state above our generated color images are not learned to be the ground-truth. So the 5$\%$ difference can be acceptable. More importantly, the proposed colorized images outperform the gray images by 2$\%$, which well supports the importance of colors on semantic understanding.

As for the joint bilateral upsampling, one may be concerned about the resolution problem. Color maps are not dominant on the resolution of the color images but the lightness channel is. A rough comparison of the peak signal-to-noise ratio (PSNR) on the images under three conditions is shown in Table~\ref{tab:table.1}. In the table, we list the means of the PSNR over the generated images without semantic information or JBU, only with semantic information, and with semantic information and JBU. Obviously, the three different settings have very similar PSNRs. The joint bilateral upsampling does not affect the quality so much but helps to preserve edge colors.

\begin{figtab}[!htb]
	\setlength{\belowcaptionskip}{-4mm}
	\begin{minipage}[b]{.45\linewidth}
		\centering
		\includegraphics[width=1\linewidth]{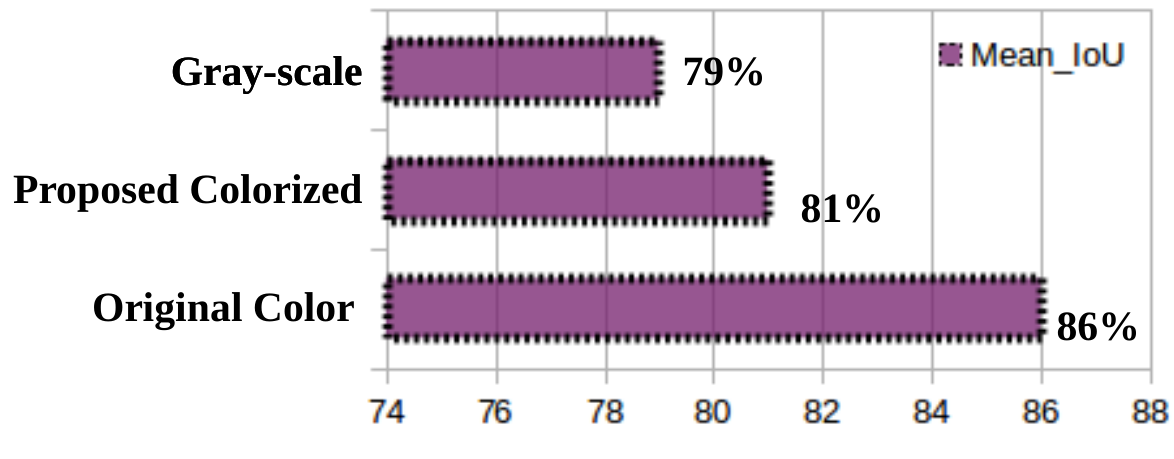}
		\vspace{-4mm}
		\figcaption [small]{Segmentation results in terms of Mean IoU of gray, proposed colorized  and original color images on PASCAL VOC2012 validation dataset. Color aids semantic segmentation.}
		\label{fig: Fig.3}
	\end{minipage}\quad
	\begin{minipage}[b]{.5\linewidth}
		\centering
		
		\makeatletter
		\def\hlinewd#1{%
			\noalign{\ifnum0=`}\fi\hrule \@height #1 \futurelet
			\reserved@a\@xhline}
		\makeatother
		
		\renewcommand\arraystretch{1.5}
		\begin{tabular}{|l|c|}
			\hlinewd{1.5pt}
			Method & PSNR    \\
			\hline\hline
			Without Semantics or JBU & 22.7    \\
			Only With Semantics & 22.3    \\
			With Semantics \& JBU & 22.0  \\
			\hlinewd{1pt}
			
		\end{tabular}
		\vspace{4mm}	
		\tabcaption[small]{Similar Mean of PSNR under three different settings on PASCAL VOC2012 validation dataset. Joint bilateral upsampling does not affect the image quality so much.}
		\label{tab:table.1}
		
	\end{minipage}	
\end{figtab}

\subsection{Ablation Study}

\begin{figure}[!htb]
	
	\begin{center}
		\includegraphics[width=1\linewidth]{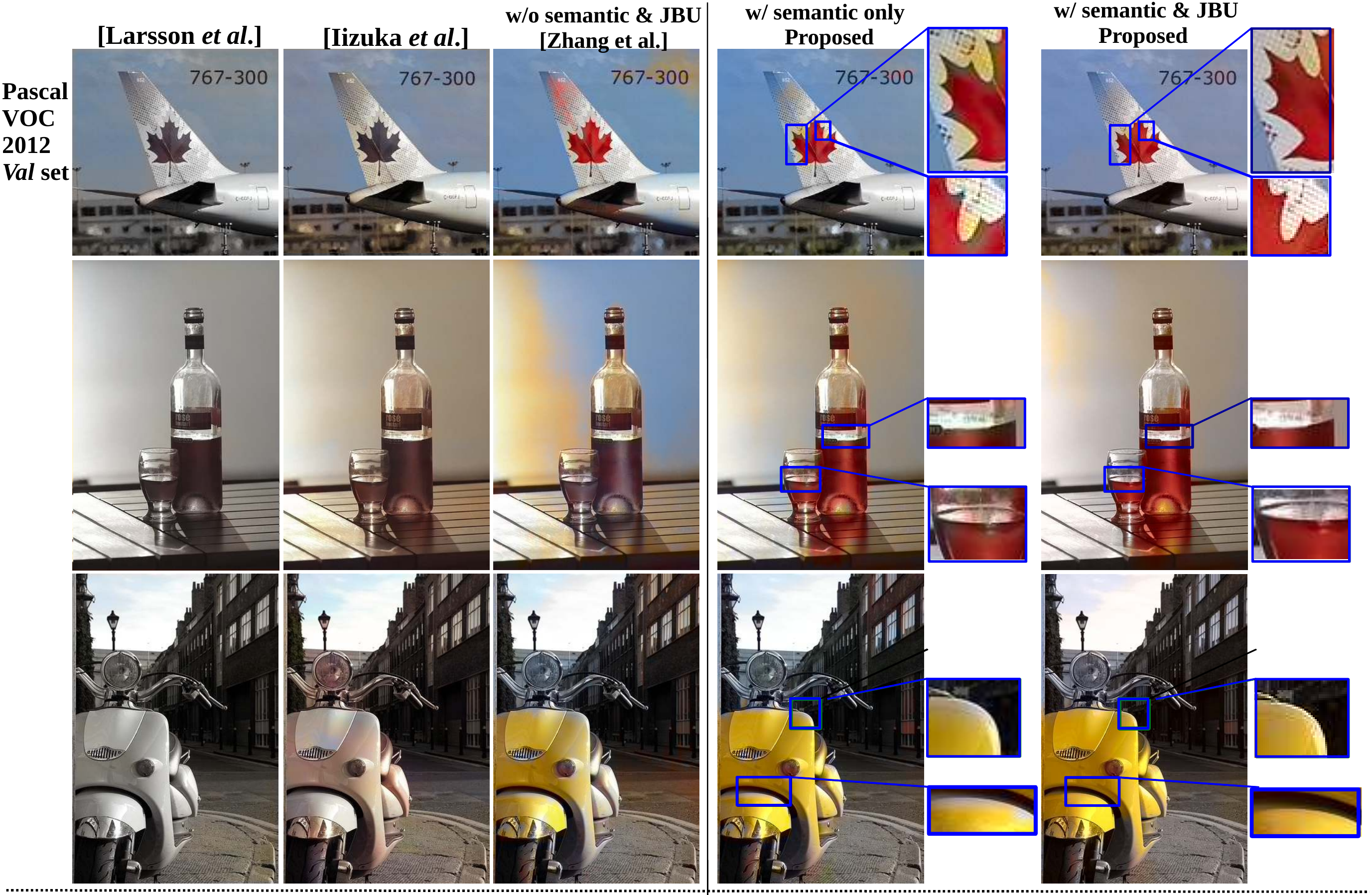}
		\includegraphics[width=1\linewidth]{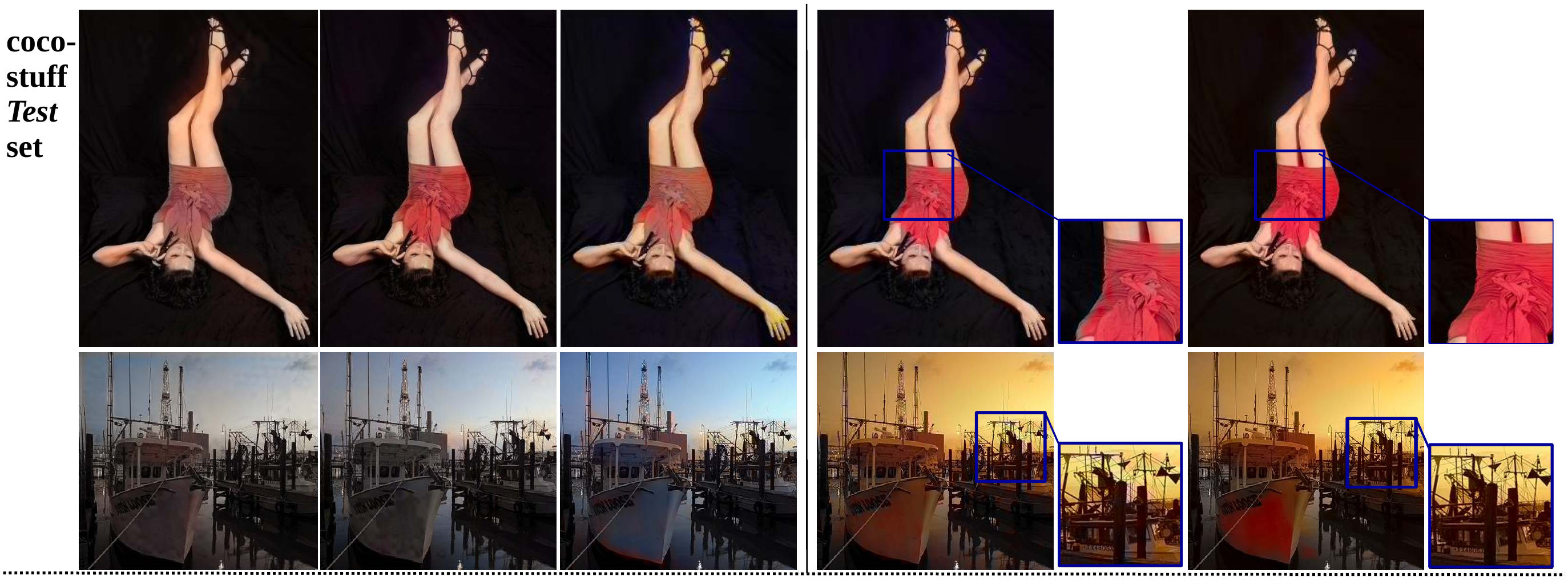}
	\end{center}
	
	\caption [small]{Example colorization results comparing the proposed methods with the state-of-the-art on Pascal VOC 2012 validation and COCO-stuff test datasets. Where the state-of-the-art suffers from desaturation, color pollution and edge bleeding, our proposed methods, with semantic priors, have better content consistence. Furthermore, with JBU, the edge colors are preserved well. We also show some local parts for detailed comparison between the edges. More results generated by our model are shown in supplementary material.}
	\label{fig: Fig.5}  	
\end{figure}

We compare our proposed methods in two settings: (1) only using semantics and (2) using semantics and JBU, with the state-of-the-art. Some successful cases on the two datasets are shown in Figure~\ref{fig: Fig.5}. The first three rows are from the PASCAL VOC2012 validation dataset and next two rows are from the COCO-stuff. As shown in the figure, the results from~\cite{Larsson2016,Iizuka2016} look grayish because colorization is treated as a regression problem in the two pipelines. In the first row, the maple leaves are not assigned correct colors by ~\cite{Larsson2016,Iizuka2016}. The sky and the tail are polluted in~\cite{Zhang2016}.  In the fourth row, the edges of the skirt are not sharp in the first three columns. The results from~\cite{Zhang2016} are more saturated but suffer from edge bleeding and color pollution. However, by  injecting semantics our methods result in better perception of the object colors. To emphasize the effect of JBU, we zoom in on local areas of the results before and after JBU. One can clearly observe the details of the edge colors. JBU helps to achieve the finer results such as the sharp edges of the maple leaf in the first row and the clear edge of the skirt in the fourth row.

\begin{figure}[!htb]
	
	\centering
	\includegraphics[width=0.9\linewidth]{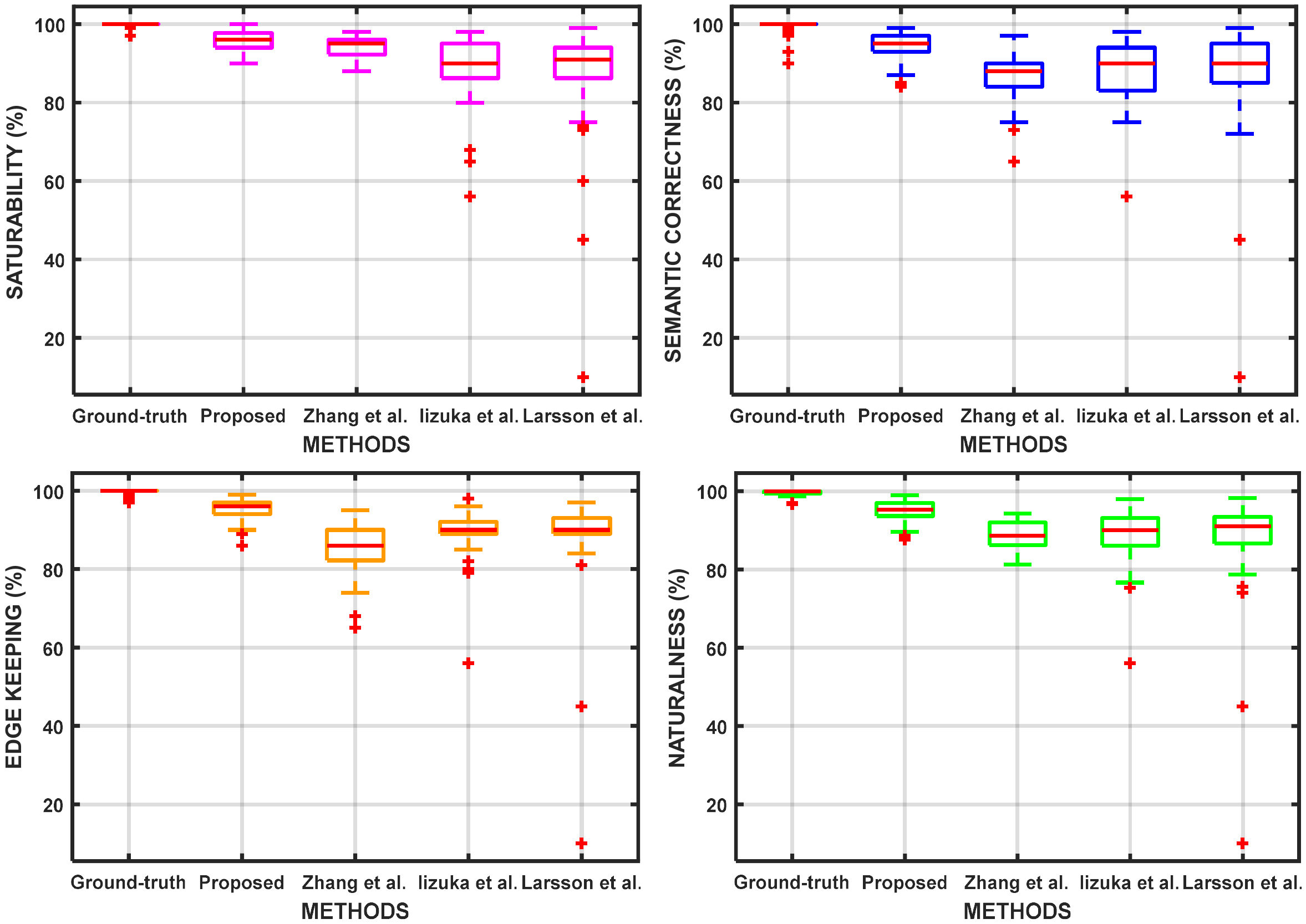}

	\caption [small]{Comparison of (a) Saturability (b) Semantic Correctness (c) Edge Keeping  and (d) Overall Naturalness. Our proposed method gets better performance in all criteria.}
	\label{fig: Fig.4}
	
\end{figure}

\subsection{Comparisons with State-of-the-art}

\begin{figure*}[!htb]

	\begin{center}
		\includegraphics[width=0.95\linewidth]{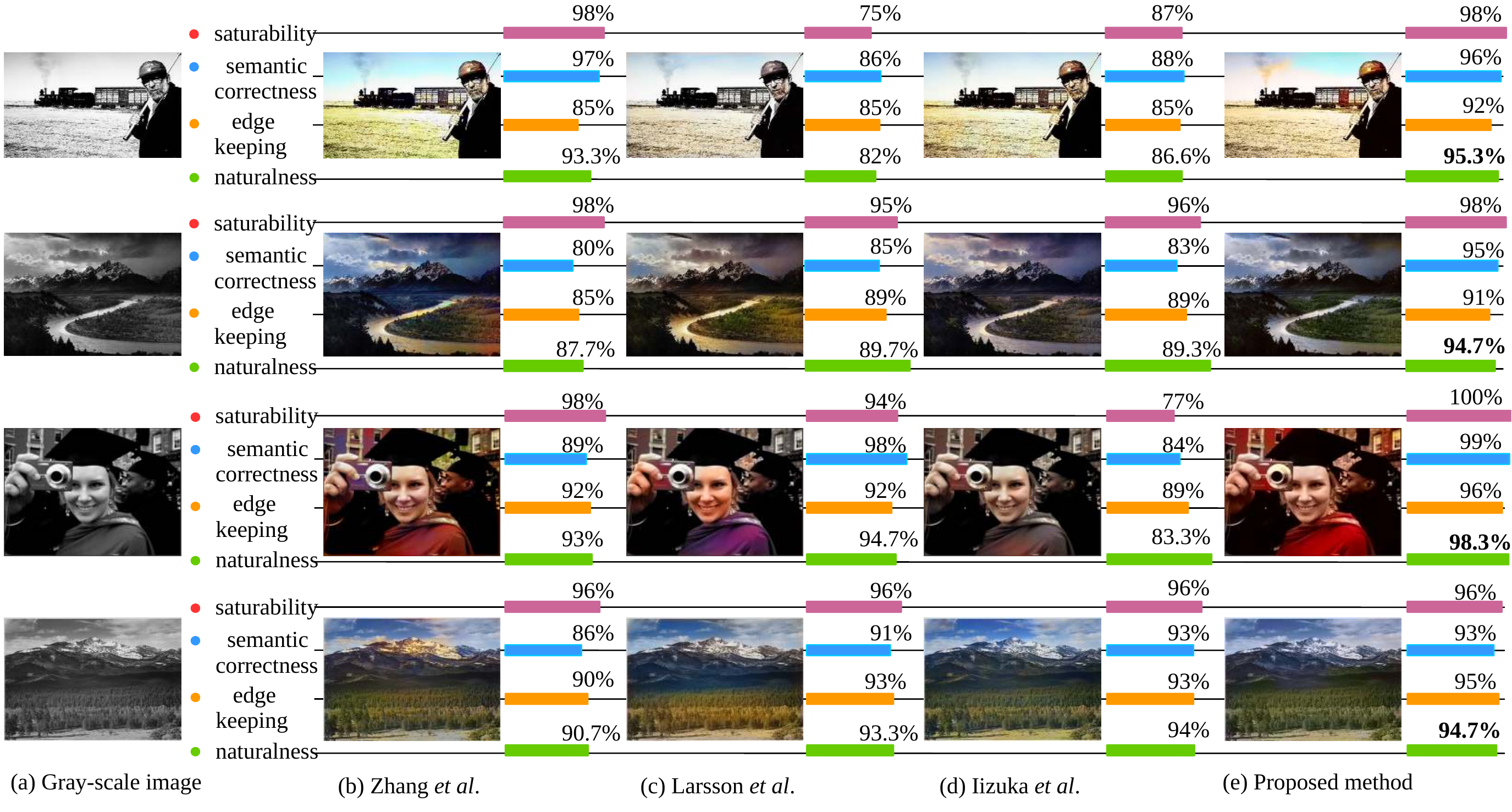}
	\end{center}

	\caption [small]{Exemplar comparison of Naturalness (includes Saturability, Semantic Correctness and Edge Keeping) with the state-of-the-art automatic colorization methods. (a) gray-scale image; (b) ~\cite{Zhang2016}; (c) ~\cite{Larsson2016}; (d) ~\cite{Iizuka2016}; (e) proposed method. Our method can produce more plausible and finer results. }
	\label{fig: Fig.6}
	
\end{figure*}

Generally, we want to produce visually compelling results which can fool a human observer, rather than recover the ground-truth. Quantitative colorization metrics may penalize reasonable, but different with ground-truth colors, especially for some artifacts (\textit{e.g.} red air balloon or green air balloon). As discussed above, colorization is a subjective issue. So qualitative results are even more important than quantitative. Similar with most papers including~\cite{Zhang2016,Iizuka2016}, we ask 20 human observers to do a real test on a combined dataset including the PASCAL VOC2012 validation and the COCO-stuff subset. Given a color image produced by our method or the three compared methods~\cite{Zhang2016,Iizuka2016,Larsson2016} or the real ground-truth image, the observers should decide whether it looks natural or not. We propose three metrics including saturability, semantic correctness and edge keeping for evaluating the naturalness. The overall naturalness is the equally weighted sum of the three values. Images are randomly selected and shown one-by-one in a few seconds to each observer. Finally, we calculate the percentage of the four criteria for each image and draw the error bar figures for comparing the images generated by different approaches (shown in Figure~\ref{fig: Fig.4}). The method in~\cite{Zhang2016} can produce rich colors but always with bad edges. The method in~\cite{Iizuka2016} keeps clear edges. Our proposed method performs better and is closer to the ground-truth. Table~\ref{tab:table.2} shows the means of the four criterion percentages of each approach. Our automatic colorization method outperforms the others considerably. We present some examples in Figure~\ref{fig: Fig.6} and label the four criterion values. For a fair comparison, the images in the last three rows are generated by the three references. It means they are taken as successful cases respectively in the three references. However, the results from~\cite{Larsson2016,Iizuka2016} look grayish in the first row. In the second row, the colors of the sky and the river from the state-of-the-art seem abnormal. In the third row, the building from~\cite{Zhang2016} is polluted and the buildings from~\cite{Larsson2016,Iizuka2016} are desaturated. Overall, our results look more realistic and saturated.

\begin{table}[!htb]	
	\setlength{\belowcaptionskip}{-6mm}	
	\makeatletter
	\def\hlinewd#1{%
		\noalign{\ifnum0=`}\fi\hrule \@height #1 \futurelet
		\reserved@a\@xhline}
	\makeatother
	\setlength{\tabcolsep}{4pt}
	\renewcommand\arraystretch{1.5}
	\begin{center}
		\begin{tabular}{|c|c|c|c|c|}
			\hlinewd{2pt}
			
			{\textbf{Method}} & \tabincell{c}{\textbf{Saturability} \\ \textbf{$(\%)$}} & \tabincell{c}{\textbf{Semantic}\\ \textbf{ Correctness $(\%)$}} & \tabincell{c}{\textbf{Edge}\\ \textbf{ Keeping $(\%)$}} & \tabincell{c}{\textbf{Naturalness}\\ \textbf{$(\%)$}}  
			\\
			\hline\hline
			\cline{1-5}	
			Iizuka \it{et al.}  & 89.00 & 87.90 & 88.90 & 88.61     \\
			\hline
			Larsson \it{et al.} & 86.00 & 86.80 & 88.00 & 86.99  \\
			\hline
			Zhang \it{et al.} & 94.00 & 86.50 & 85.40 & 88.66    \\
			\hline
			This Paper  & \textbf{95.70} & \textbf{94.10} & \textbf{94.80} & \textbf{94.89}  \\
			\hline\hline
			Ground-truth & 99.69 & 99.27 & 99.76 & 99.58  \\
			\hlinewd{2pt}
			
		\end{tabular}
	\end{center}
	\caption{Comparison of Naturalness between the state-of-the-art and the proposed methodon the combined dataset. The mean value of each criterion is shown. Our proposed method has better performance.}
	\label{tab:table.2}
\end{table}

\subsection{Failure Cases} 
\begin{figure}[!htb]
	
	\centering
	\includegraphics[width=0.95\linewidth]{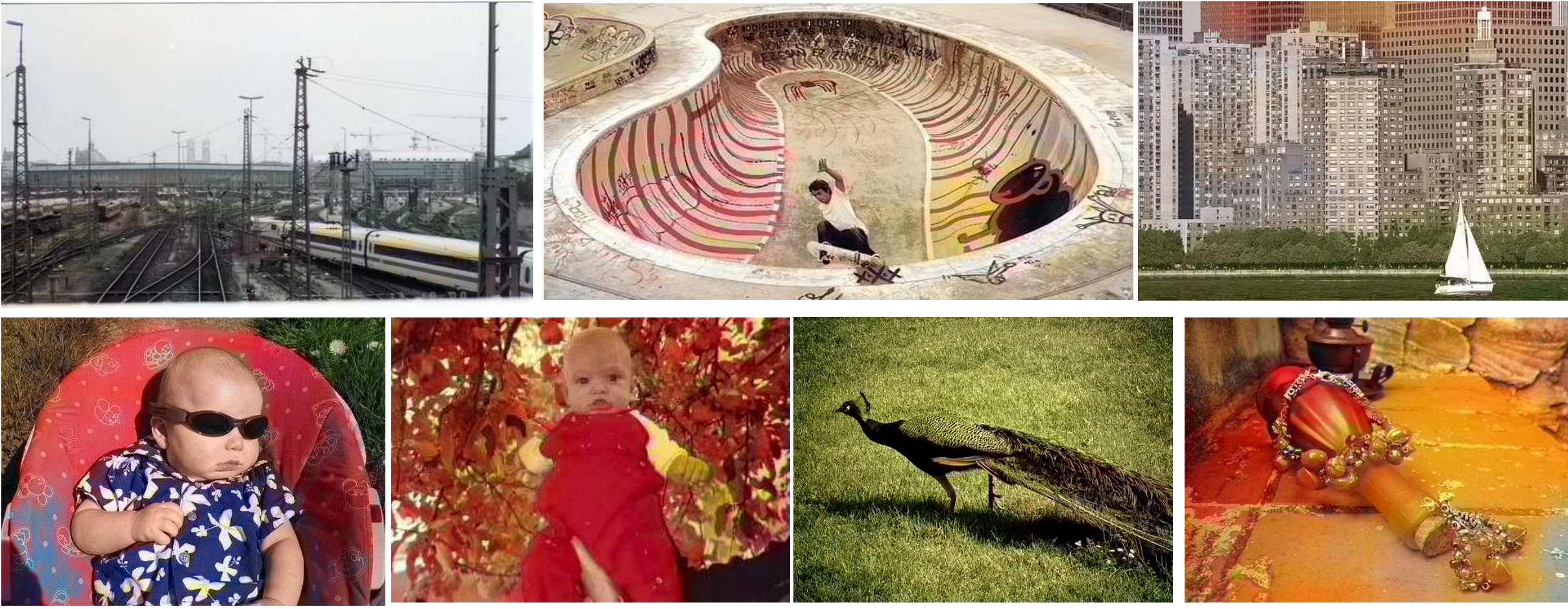}
	
	\caption{Failure cases. \textit{Top row, left-to-right}: not enough colors; incorrect semantic understanding; \textit{bottom row, left-to-right}: background inconsistence; small object confusion (leaves and apples); lack of object categories (peacock, jewelry). }
	\label{fig: Fig.7}		
\end{figure}

Our method can output plausible colorized images but it is not perfect. There are still some common issues encountered by the proposed approach and also other automatic systems. We provide a few failure cases in Figure~\ref{fig: Fig.7}. It is believed that incorrect semantic understanding results in unreasonable colors. Though we incorporate semantics for improving colorization, there are not enough categories. We assume a finer semantic segmentation with more class labels will further enhance the results.

\section{Conclusion}
\label{sec:concl}

In this paper, we address two general problems of current automatic colorization approaches: context confusion and edge color bleeding. Our hierarchical structure with semantic segmentation and colorization was designed to strengthen the ability of semantic understanding so that content confusion will be reduced. And our joint bilateral upsampling layer successfully preserves edge colors at inference. We achieved satisfying results in most cases. Our code will be released to foster further improvements in the future.

\bibliography{egbib}

\begin{thebibliography}{31}
\providecommand{\natexlab}[1]{#1}
\providecommand{\url}[1]{\texttt{#1}}
\expandafter\ifx\csname urlstyle\endcsname\relax
  \providecommand{\doi}[1]{doi: #1}\else
  \providecommand{\doi}{doi: \begingroup \urlstyle{rm}\Url}\fi

\bibitem[Baig and Torresani(2017)]{Baig2017}
Mohammad~Haris Baig and Lorenzo Torresani.
\newblock Multiple hypothesis colorization and its application to image
  compression.
\newblock \emph{Computer Vision and Image Understanding}, 164:\penalty0
  111--123, 2017.

\bibitem[Caesar et~al.(2016)Caesar, Uijlings, and Ferrari]{Ca2016}
Holger Caesar, Jasper Uijlings, and Vittorio Ferrari.
\newblock Coco-stuff: Thing and stuff classes in context.
\newblock \emph{arXiv:1612.03716}, 2016.

\bibitem[Cao et~al.(2017)Cao, Zhou, Zhang, and Yu]{Cao2017}
Yun Cao, Zhiming Zhou, Weinan Zhang, and Yong Yu.
\newblock Unsupervised diverse colorization via generative adversarial
  networks.
\newblock \emph{arXiv:1702.06674}, 2017.

\bibitem[Charpiat et~al.(2008)Charpiat, Hofmann, and Schölkopf]{Charpiat2008}
Guillaume Charpiat, Matthias Hofmann, and Bernhard Schölkopf.
\newblock Automatic image colorization via multimodal predictions.
\newblock In \emph{European Conference on Computer Vision}, pages 126--139,
  2008.

\bibitem[Chen et~al.(2016)Chen, Papandreou, Kokkinos, Murphy, and
  Yuille]{Chen16}
Liang-Chieh Chen, George Papandreou, Iasonas Kokkinos, Kevin Murphy, and
  Alan~L. Yuille.
\newblock Deeplab: Semantic image segmentation with deep convolutional nets,
  atrous convolution, and fully connected crfs.
\newblock \emph{IEEE Transactions on Pattern Analysis and Machine
  Intelligence}, 2016.

\bibitem[Cheng et~al.(2015)Cheng, Yang, and Sheng]{Cheng2015}
Zezhou Cheng, Qingxiong Yang, and Bin Sheng.
\newblock Deep colorization.
\newblock In \emph{{IEEE} International Conference on Computer Vision}, 2015.

\bibitem[Chia et~al.(2011)Chia, Zhuo, Gupta, and Tai]{Chia2011}
Alex Yong-Sang Chia, Shaojie Zhuo, Raj~Kumar Gupta, and Yu-Wing Tai.
\newblock Semantic colorization with internet images.
\newblock \emph{ACM Transactions on Graphics (TOG)}, 30\penalty0 (6), 2011.

\bibitem[Dai et~al.(2016)Dai, Li, He, and Sun]{Jifeng2016}
Jifeng Dai, Yi~Li, Kaiming He, and Jian Sun.
\newblock R-fcn: Object detection via region-based fully convolutional
  networks.
\newblock In \emph{Neural Information Processing Systems(NIPS)}, 2016.

\bibitem[Everingham et~al.(2012)Everingham, Gool, Williams, Winn, and
  Zisserman]{pascal2012}
Mark Everingham, Luc~Van Gool, Christopher~K.L. Williams, John Winn, and Andrew
  Zisserman.
\newblock The pascal visual object classes challenge 2012 (voc2012) results.
\newblock
  http://www.pascal-network.org/challenges/VOC/voc2012/workshop/index.html,
  2012.

\bibitem[Frans(2017)]{Frans2017}
Kevin Frans.
\newblock Outline colorization through tandem adversarial networks.
\newblock \emph{arXiv:1704.08834}, 2017.

\bibitem[Guadarrama et~al.(2017)Guadarrama, Dahl, Bieber, Norouzi, Shlens, and
  Murphy]{Gua2017}
Sergio Guadarrama, Ryan Dahl, David Bieber, Mohammad Norouzi, Jonathon Shlens,
  and Kevin Murphy.
\newblock Pixcolor: Pixel recursive colorization.
\newblock \emph{arXiv:1705.07208}, 2017.

\bibitem[Guo et~al.(2017)Guo, Pan, Lei, and Ding]{Guo2017}
Jiayi Guo, Zongxu Pan, Bin Lei, and Chibiao Ding.
\newblock Automatic color correction for multisource remote sensing images with
  wasserstein cnn.
\newblock \emph{Remote Sensing}, 9:\penalty0 483, 2017.

\bibitem[Gupta et~al.(2012)Gupta, Chia, Rajan, Ng, and Huang]{Gupta2012}
Raj~Kumar Gupta, Alex Yong-Sang Chia, Deepu Rajan, Ee~Sin Ng, and Zhiyong
  Huang.
\newblock Image colorization using similar images.
\newblock In \emph{ACM international conference on Multimedia}, pages 369--378,
  2012.

\bibitem[Huang et~al.(2005)Huang, Tung, Chen, Wang, and Wu]{Huang2005}
Yi-Chin Huang, Yi-Shin Tung, Jun-Cheng Chen, Sung-Wen Wang, and Ja-Ling Wu.
\newblock An adaptive edge detection based colorization algorithm and its
  applications.
\newblock In \emph{ACM international conference on Multimedia}, pages 351--354,
  2005.

\bibitem[Iizuka et~al.(2016)Iizuka, Simo-Serra, and Ishikawa]{Iizuka2016}
Satoshi Iizuka, Edgar Simo-Serra, and Hiroshi Ishikawa.
\newblock Let there be color!: Joint end-to-end learning of global and local
  image priors for automatic image colorization with simultaneous
  classification.
\newblock In \emph{Conference on Special Interest Group on Computer GRAPHics
  and Interactive Techniques}, 2016.

\bibitem[Irony et~al.(2005)Irony, Cohen-Or, and Lischinski]{Irony2005}
Revital Irony, Daniel Cohen-Or, and Dani Lischinski.
\newblock Colorization by example.
\newblock In \emph{EGSR '05 Proceedings of the Sixteenth Eurographics
  conference on Rendering Techniques}, pages 201--210, 2005.

\bibitem[Isola et~al.(2017)Isola, Zhu, Zhou, and Efros]{Isola2017}
Phillip Isola, Jun-Yan Zhu, Tinghui Zhou, and Alexei~A. Efros.
\newblock Image-to-image translation with conditional adversarial networks.
\newblock In \emph{{IEEE} Conference on Computer Vision and Patten
  Recognition}, 2017.

\bibitem[Kopf et~al.(2007)Kopf, Cohen, Lischinski, and Uyttendaele]{KCLU07}
Johannes Kopf, Michael~F. Cohen, Dani Lischinski, and Matt Uyttendaele.
\newblock Joint bilateral upsampling.
\newblock \emph{ACM Transactions on Graphics (Proceedings of SIGGRAPH 2007)},
  26\penalty0 (3), 2007.

\bibitem[Larsson et~al.(2016)Larsson, Maire, and Shakhnarovich]{Larsson2016}
Gustav Larsson, Michael Maire, and Gregory Shakhnarovich.
\newblock Learning representations for automatic colorization.
\newblock In \emph{European Conference on Computer Vision}, 2016.

\bibitem[Levin et~al.(2004)Levin, Lischinski, and Weiss]{Levin2004}
Anat Levin, Dani Lischinski, and Yair Weiss.
\newblock Colorization using optimization.
\newblock \emph{ACM Transactions on Graphics (TOG) (Proceedings of ACM SIGGRAPH
  2004)}, 23:\penalty0 689--694, 2004.

\bibitem[Limmer and Lensch(2016)]{Limmer2016}
Matthias Limmer and Hendrik~P.A. Lensch.
\newblock Infrared colorization using deep convolutional neural networks.
\newblock In \emph{IEEE International Conference on Machine Learning and
  Applications}, 2016.

\bibitem[Lin et~al.(2014)Lin, Maire, Belongie, Bourdev, Girshick, Hays, Perona,
  Ramanan, Zitnick, and Dollár]{Lin2014}
Tsung-Yi Lin, Michael Maire, Serge Belongie, Lubomir Bourdev, Ross Girshick,
  James Hays, Pietro Perona, Deva Ramanan, C.~Lawrence Zitnick, and Piotr
  Dollár.
\newblock Microsoft coco: Common objects in context.
\newblock \emph{arXiv:1405.0312}, 2014.

\bibitem[Liu et~al.(2008)Liu, Wan, Qu, Wong, Lin, Leung, and Heng]{Liu2008}
Xiaopei Liu, Liang Wan, Yingge Qu, Tien-Tsin Wong, Stephen Lin, Chi-Sing Leung,
  and Pheng-Ann Heng.
\newblock Intrinsic colorization.
\newblock \emph{ACM Transactions on Graphics (TOG) (Proceedings of ACM SIGGRAPH
  Asia 2008)}, 2008.

\bibitem[Luan et~al.(2007)Luan, Wen, Cohen-Or, Liang, Xu, and Shum]{Luan2007}
Qing Luan, Fang Wen, Daniel Cohen-Or, Lin Liang, Ying-Qing Xu, and Heung-Yeung
  Shum.
\newblock Natural image colorization.
\newblock In \emph{EGSR'07 Proceedings of the 18th Eurographics conference on
  Rendering Techniques}, pages 309--320, 2007.

\bibitem[Morimotoand et~al.(2009)Morimotoand, Taguchii, and Naemura]{Mor2009}
Yuji Morimotoand, Yuichi Taguchii, and Takeshi Naemura.
\newblock Automatic colorization of grayscale images using multiple images on
  the web.
\newblock In \emph{{ACM} Conference on Special Interest Group on Computer
  GRAPHics and Interactive Techniques}, 2009.

\bibitem[Noh et~al.(2015)Noh, Hong, and Han]{Noh2015}
Hyeonwoo Noh, Seunghoon Hong, and Bohyung Han.
\newblock Learning deconvolution network for semantic segmentation.
\newblock In \emph{{IEEE} International Conference on Computer Vision}, 2015.

\bibitem[Qu et~al.(2006)Qu, Wong, and Heng]{Qu2006}
Yingge Qu, Tien-Tsin Wong, and Pheng-Ann Heng.
\newblock Manga colorization.
\newblock In \emph{{ACM} Conference on Special Interest Group on Computer
  GRAPHics and Interactive Techniques}, 2006.

\bibitem[Royer et~al.(2017)Royer, Kolesnikov, and Lampert]{Royer2017}
Amelie Royer, Alexander Kolesnikov, and Christoph~H. Lampert.
\newblock Probabilistic image colorization.
\newblock In \emph{British Machine Vision Conference}, 2017.

\bibitem[Shelhamer et~al.(2016)Shelhamer, Long, and Darrell]{She2016}
Evan Shelhamer, Jonathan Long, and Trevor Darrell.
\newblock Fully convolutional networks for semantic segmentation.
\newblock \emph{IEEE Transactions on Pattern Analysis and Machine
  Intelligence}, 39:\penalty0 640--651, 2016.

\bibitem[Zhang et~al.(2016)Zhang, Isola, and Efros]{Zhang2016}
Richard Zhang, Phillip Isola, and Alexei~A. Efros.
\newblock Colorful image colorization.
\newblock In \emph{European Conference on Computer Vision}, 2016.

\bibitem[Zhang et~al.(2017)Zhang, Zhu, Isola, Geng, S.Lin, and
  Efros]{Zhang2017}
Richard Zhang, Jun-Yan Zhu, Phillip Isola, Xinyang Geng, Angela S.Lin, and
  Alexei~A. Efros.
\newblock Real-time user-guided image colorization with learned deep priors.
\newblock In \emph{Conference on Special Interest Group on Computer GRAPHics
  and Interactive Techniques}, 2017.

\end{thebibliography}
\end{document}